%% file: main.tex
\documentclass[letterpaper]{article} % DO NOT CHANGE THIS
\usepackage{aaai25}  % DO NOT CHANGE THIS <- [submission]{aaai25} to reveal names
\usepackage{times}  % DO NOT CHANGE THIS
\usepackage{helvet}  % DO NOT CHANGE THIS
\usepackage{courier}  % DO NOT CHANGE THIS
\usepackage[hyphens]{url}  % DO NOT CHANGE THIS
\usepackage{graphicx} % DO NOT CHANGE THIS
\usepackage{natbib}  % DO NOT CHANGE THIS AND DO NOT ADD ANY OPTIONS TO IT
\usepackage{caption} % DO NOT CHANGE THIS AND DO NOT ADD ANY OPTIONS TO IT
\usepackage{algorithm}
\usepackage{algpseudocode}
\usepackage{mathtools}
\usepackage[export]{adjustbox}
\usepackage{amssymb}
\usepackage{amsthm}
\usepackage{booktabs}
\usepackage{algpseudocode}
\usepackage{algorithm}
\usepackage{dsfont}
\usepackage{arydshln}

\usepackage{nameref}

\usepackage{newfloat}
\usepackage{listings}
\usepackage{bibentry}
\DeclareCaptionStyle{ruled}{labelfont=normalfont,labelsep=colon,strut=off} % DO NOT CHANGE THIS
\floatstyle{ruled}
\newfloat{listing}{tb}{lst}{}
\floatname{listing}{Listing}
\title{Meta-TTT: A Meta-learning Minimax Framework For Test-Time Training}
\author {
    Chen Tao, %\textsuperscript
    Li Shen, %\textsuperscript{\rm 1,2}
    Soumik Mondal %\textsuperscript
}
\affiliations {
    Institute for Infocomm Research (I2R), Agency for Science, Technology and Research (A*STAR)\\
    taoc@i2r.a-star.edu.sg, lshen@i2r.a-star.edu.sg, soumik\_mondal@i2r.a-star.edu.sg
}

\begin{document}

\maketitle
\input{sec/0_abstract}    
\input{sec/1_intro}

\input{sec/2_related_work}
\input{sec/3_method}
\input{sec/4_experiments}
\input{sec/5_conclusion}


\appendix

\bibliography{main}
\input{sec/X_suppl}
\end{document}

%% file: sec/0_abstract.tex
\begin{abstract}
Test-time domain adaptation is a challenging task that aims to adapt a pre-trained model to limited, unlabeled target data during inference. Current methods that rely on self-supervision and entropy minimization underperform when the self-supervised learning (SSL) task does not align well with the primary objective. Additionally, minimizing entropy can lead to suboptimal solutions when there is limited diversity within minibatches. This paper introduces a meta-learning minimax framework for test-time training on batch normalization (BN) layers, ensuring that the SSL task aligns with the primary task while addressing minibatch overfitting. We adopt a \textit{mixed-BN} approach that interpolates current test batch statistics with the statistics from source domains and propose a stochastic domain synthesizing method to improve model generalization and robustness to domain shifts. Extensive experiments demonstrate that our method surpasses state-of-the-art techniques across various domain adaptation and generalization benchmarks, significantly enhancing the pre-trained model's robustness on unseen domains. 
\end{abstract}

%% file: sec/1_intro.tex
\section{Introduction}
\label{sec:intro}

Deep neural networks are highly effective in learning from data but operate under the assumption that the distributions of training and test data are identical. In real-world scenarios, challenges in data collection can result in domain shifts, where the characteristics of the test data differ from those of the training data. This mismatch often leads to decreased performance on the test dataset.

The idea of domain generalization (DG) is to learn a domain-agnostic model which can be applied to an unseen domain. A recent study~\cite{gulrajani2021in} highlights the challenges many existing methods face. Specifically, their underperformance compared to the baseline empirical risk minimization (ERM) method is primarily due to the focus on addressing distribution shifts in the training data while neglecting test samples. Domain adaptation (DA) methods address this problem by aligning the distributions of training and test samples~\cite{hoffman2018cycada, sun2016Coral, luo2019taking, kang2019contrastive, saito2018maximum, saito2019semi}, but most of these methods rely on the co-existence of source and target data or at least a part of the source information. Consequently, source-free unsupervised domain adaptation aims to solve the domain shift problem by unlabeled target data alone~\cite{liang2020SHOT, yang2020iccv, li2020cvpr}. However, their approach requires high computational costs and the access to an entire target domain during test time, whereas in many real world scenarios, only a narrow portion of target data is available.

Recently, test-time adaptation (TTA) has emerged as a solution to the challenging task of adapting a pre-trained model to a new test distribution by learning from unlabeled test (target) data during inference. Given the limited availability of unlabeled target data, the common adaptation approach involves exploiting the information from unseen samples by fine-tuning parts of the model, such as the BN layers, in an unsupervised manner. The training objective typically employs entropy minimization or similar SSL losses, which are both simple and computationally efficient~\cite{li2018AdaBN, wang2021tent, zhang2023adapt, lim2023ttn,niu2023}. However, the success of this adaptation method relies on accurately estimating BN statistics, a task that is difficult to achieve with limited unseen data and a large domain gap. Consecutive updates to the adaptable weights on narrow portions of the target distribution can lead to model overfitting. Moreover, in scenarios with low intra-batch diversity, entropy minimization can predict the same class or a very limited number of classes in most cases. Recent works \cite{you2021alpha, lim2023ttn, zhang2023adapt} have proposed to mix statistics calculated from the current test batch and the running statistics obtained from source data by adjusting the interpolation weight $\alpha$. For simplicity, we refer to this approach as \textit{mixed-BN} throughout this paper. Despite the soundness of mixing batch statistics, a sub-optimal choice of hyperparameters can lead to catastrophic failure. While generalizing these hyperparameters across unforeseen domains is desirable, it remains a significant challenge.

Recent studies of test-time training (TTT) for domain shifts have shown promising results \cite{you2021alpha, liu2021ttt++, Chen2023TTA, jang2022nearest}. Nonetheless, TTT methods carry a risk of significantly degrading the model's performance on the main task during adaptation through a SSL task, leading to unexpected failures and lower test accuracy. This problem is largely attributed to unconstrained model updates from the SSL task that interferes with the main task. Liu \cite{liu2021ttt++} highlights the importance of incorporating a SSL task highly correlated with the main task for test-time training. 

In this paper, we present a novel meta-learning framework for \textit{mixed-BN} based test-time training (Meta-TTT), aimed at aligning the test-time SSL task with the primary task while mitigating minibatch overfitting. Our meta-training process learns hyperparameters tailored to the \textit{mixed-BN} model. We also enhance the model's robustness by exposing it to various augmented domain shifts during training, and this is achieved through a stochastic simulation method utilizing a linear transformation layer. Additionally, we propose a minimax entropy approach to adversarially optimize the \textit{mixed-BN} model, boosting its resilience in unseen test streams and preventing oversimplified solutions. Our contributions are summarized blow:
\begin{itemize}
  \item Propose a meta-learning framework to align the SSL task with the main task and learn the hyperparameters of the \textit{mixed-BN} model with good generalization capabilities.  
  \item Propose a minimax entropy for the SSL task to prevent suboptimal solutions caused by entropy minimization. 
  \item Introduce a stochastic method to synthesize diverse domain shift in each batch.
\end{itemize}

%% file: sec/2_related_work.tex
\section{Related Work}

\subsection{Unsupervised Domain Adaptation}

Unsupervised domain adaptation (UDA) methods align the distributions of training and test samples in the input \cite{hoffman2018cycada}, feature \cite{sun2016Coral, saito2018maximum, kang2019contrastive}, and output spaces \cite{luo2019taking} by minimizing a divergence measure \cite{sun2016Coral, saito2019semi} or using a SSL loss \cite{hoffman2018cycada, kang2019contrastive, luo2019taking}. Some intrinsic limitations in this line of work include:
\begin{itemize}
    \item {Dependence on the Co-existence of Source and Target Data:} UDA methods depend on the co-existence of source and target data, which can be impractical in real-world scenarios due to data privacy concerns.
    \item {Requirements and Assumptions of Source-free Domain Adaptation Approaches:} To address the co-existence of source and target data, source-free domain adaptation approaches \cite{liang2020SHOT, li2020cvpr, yang2020iccv} have been proposed, which updates models solely on unlabeled target data during testing. However, these approaches require access to the entire target data and assume a stationary domain during training, which limits their practicality. 
\end{itemize}
In contrast, our method emphasizes online adaptation, where each target sample is processed by the model only once. Additionally, our approach does not require source data during testing, making it more practical and mitigating the above limitations.

\subsubsection{Adversarial optimization of domain divergence} 
Saito \cite{saito2018maximum} maximized the discrepancy between the outputs of two classifiers followed by minimizing the feature discrepancy of a feature extractor in an adversarial manner to achieve feature alignment. Building on this, Saito \cite{saito2019semi} introduced a minimax entropy approach for semi-supervised domain adaptation, which alternates between maximizing the conditional entropy of unlabeled target data to update the classifier and minimizing it to update the feature encoder, thereby adversarially optimizing the model. Inspired by these works, we propose a minimax entropy approach to optimize the domain divergence for test-time \textit{mixed-BN} adaptation.

\subsection{Domain Generalization}
Domain generalization (DG) typically involves training models exclusively on source domains to generalize to unseen target domains. Various methods, including adversarial training \cite{li2018CVPR, deng2020representation} and invariant representation learning \cite{shi2021gradient, rame2022fishr}, aim to learn domain-agnostic features. Data augmentation \cite{zhou2021mixstyle, nam2021reducing,yan2020improve,li2021simple} has also been proposed to increase the diversity of training data for generalization tasks. In the paper, we introduce a stochastic augmentation approach to simulate distribution shifts for batch data. It is important to note that a fundamental difference between domain generalization and our setup is the lack of utilization of test-time statistics in DG.

\subsubsection{Meta-learning For Domain Generalization}
The goal of meta-learning is to train a model across diverse learning tasks using an episodic training strategy, enabling it to handle new tasks efficiently with a few training samples. Finn ~\cite{finn2017MAML} introduced a Model-Agnostic Meta-Learning algorithm (MAML) for few-shot learning. Building on this, Li \cite{li2018MLDG} proposed a gradient based Meta-Learning method for Domain Generalization (MLDG). Balaji ~\cite{balaji2018metareg} developed MetaReg, a Meta-Regularizer for the classifier, and Dou \cite{dou2019domain} introduced two complementary losses to explicitly regularize the semantic structure of feature space for MLDG. Zhang  proposed a framework of Adaptive Risk Minimization (ARM) \cite{zhang2021ARM}, which is closely aligned with our work. There are variations within the ARM family. For example, ARM-BN involves training a unified model to predict BN parameters using unlabeled data from unseen domains, while ARM-LL employs a loss network to generate a SSL loss, updating the model based on both the SSL loss and main loss.

In our approach, rather than directly training a SSL loss function or a model to predict BN parameters, we adopt a meta-learning paradigm. This approach aligns the SSL task with the main task and learns hyperparameters that generalize robustly across diverse domains.

\subsection{Test-Time Adaptation}

Test-time adaptation (TTA) handles distribution shifts during testing by partitioning the network into adaptable and frozen parameters and optimizing the trainable ones in a unsupervised manner usually in mini-batches~\cite{wang2021tent,Chen2023TTA, wang2022COTTA,boudiaf2022LAME}, or even at an instance level~\cite{wang2022COTTA}. 

% TTA methods employ diverse strategies for handling adaptable parameters. For example, CoTTA~\cite{wang2022COTTA} develops a stochastic restoration technique that randomly adapts a small number of parameters. Some parameter-free methods, such as Laplacian Adjusted Maximum-likelihood Estimation (LAME)~\cite{boudiaf2022LAME} and prototypical Classification (T3A)~\cite{iwasawa2021t3a}, directly update output assignment vectors.

Despite the notable success of previous methods in domain adaptation, we identify and explore several common limitations. Our work aims to address and alleviate these challenges, which include:

\begin{itemize}
  \item {Limitation of Accurate BN Statistics:} For adaptable parameters, one popular choice is the BN layers. AdaBN~\cite{li2018AdaBN} replaces source BN statistics with target statistics at test batches. Tent~\cite{wang2021tent} extends AdaBN by learning affine transformation parameters during test time. The success of these methods relies on an accurate estimation of the BN statistics. It becomes challenging with limited test data, potentially leading to model overfitting when updates are based on small segments of the target distribution.
  \item {Challenge of Determining Mixing Coefficients:} Recent methods \cite{you2021alpha, lim2023ttn, zhang2023adapt} propose the mixing of source and target statistics during test time. Yet, determining the mixing coefficients often involves manual selection as hyperparameters \cite{you2021alpha} or learning from source domains as priors \cite{lim2023ttn}. Sub-optimal choices of these hyperparameters might lead to catastrophic failures.
  \item {Issue with Entropy Minimization and Small Batch Sizes:} Entropy minimization is a widely-used unsupervised training objective in TTA \cite{wang2021tent, lim2023ttn, zhang2023adapt, niu2023}. MEMO \cite{zhang2022memo} and CoTTA \cite{wang2022COTTA} further minimized the entropy of average predictions over different augmentations. However, entropy minimization often assumes large test batch sizes. In scenarios involving smaller batch sizes and class-imbalanced data, studies \cite{boudiaf2022LAME, liang2020SHOT, zhang2022lccs} illustrated that the entropy minimization method might degrade the model, collapsing to narrow distributions such as consistent predictions for the same few classes.
\end{itemize}

\subsection{Test-Time Training}
Alternatively, and more closely related to our method, test-time training (TTT) strategies involve modifying the training objective by training the model on source data through the main task combined with a SSL task, such as rotation prediction \cite{sun2020test}, nearest source prototypes \cite{choi2022SWR, jang2022nearest}, or utilizing consistency loss \cite{Chen2023TTA}. The model is then updated based on the SSL task during test time. TTT addresses potential privacy concerns by not requiring the revisitation of source data or retention of source statistics during test time. Furthermore, TTT methods do not rely on assumptions about the output of the main task, making them more generic and applicable to a wide range of scenarios beyond classification problems, including instance tracking \cite{Fu2021LearningTT} and reinforcement learning \cite{Hansen2020SelfSupervisedPA}.

\subsubsection{Challenges In Test-Time Training}
Although TTT emerges as a promising paradigm for domain adaptation and generalization, it can possibly harm, rather than improve, the test-time performance of severe distribution shifts. A recent study \cite{liu2021ttt++} emphasized the importance of the alignment between the main task and the SSL task for achieving high test accuracy after adaptation. An uncontrolled update from the SSL loss might interfere with the primary task. The study shows that TTT can improve model performance only when the SSL loss closely coordinates with the main loss. However, achieving such alignment is particularly challenging when dealing with data from unseen domains.

%% file: sec/3_method.tex
\section{Our Method}
Let $\Theta (x)$ denotes the parameters of a model trained on the labeled source domain $D_s = \{ (x_i, y_i )\}_{i=1}^N$, where $x_i\in X$ is an input and $y_i \in Y$ is its corresponding label. During test time, we consider a sequence of unlabeled test minibatches $D_t=\{B_t\}$. Each test minibatch $B_t$ is sampled from an arbitrary unseen target distribution $p_t(x)$. We work under the standard covariant shift assumption: $p_s(y|x) = p_t(y|x)$ and $p_s(x) \neq p_t(x)$.  The goal of our method Meta-TTT is to improve performance on $B_t$ by adapting the trained model to target distribution.

We begin by explaining the \textit{mixed-BN} adaptation scheme in Section~\nameref{sec:method_BN}. In section~\nameref{sec:method}, we introduce a novel minimax entropy loss and a meta-learning framework, which optimizes both the proposed SSL loss and the main loss to ensure their coordinated descent and to facilitate the learning of the interpolating parameter $\alpha$. In section~\nameref{sec:method_aug}, we present an augmentation technique that synthesizes domain shifts within minibatches. 

\subsection{Mixed-BN Adaptation}
\label{sec:method_BN}
We adopt the channel-wise interpolated BN proposed by Singh \cite{singh2019evalnorm} and Summers \cite{summers2019four}, which is known for its resilience against error accumulation and forgetting caused by distribution shifts over time \cite{lim2023ttn}. Formally, the BN statistics are computed as: 
\begin{equation}
\begin{split}
\mu &= \alpha \mu_t + (1-\alpha) \mu_s \\
\sigma^2 &= \alpha \sigma_t^2 + (1-\alpha)\sigma_s^2 + \alpha(1-\alpha)(\mu_t - \mu_s)^2,
\end{split}
\label{eq:BN}
\end{equation}
where $\{\mu_s, \sigma_s\}$ and$\{\mu_t, \sigma_t\}$ are the source and  target batch statistics, respectively. The interpolating weight $\alpha \in \mathbb{R}^c$ falls within the range of $[0, 1]$, where $c$ is the number of feature channels. A BN layer transforms input features $z$ into $\hat{z} =\gamma (z-\mu)/\sigma + \beta$ with the affine parameter vectors $\beta$ and  $\gamma$ of size $c$. Notably, we consider $\alpha$ a learnable parameter rather than a hyperparameter, along with the affine parameters $\beta$ and $\gamma$.

During test time, we adapt parameters $\Theta_{ad} =\{\gamma_{l,c}, \beta_{l,c}, \alpha_{l,c}\}$ through a minimax entropy loss for each BN layer $l$ and channel $c$ while keeping the remaining parameters $\Theta_{fz} = \Theta \backslash \Theta_{ad}$ frozen.

\subsection{Meta-Learning for Test-time Training}
\label{sec:method}
In this section, we first introduce a minimax entropy loss for the SSL task, then describe the meta-learning paradigm used to train both the proposed SSL loss and the model's main task for TTT.

\subsubsection{Minimax Entropy Objective}
\label{sec:method_minimax}
We partition the test-time adaptable parameters $\Theta_{ad}$ into $\Theta_\beta = \{\beta_{l\in L}\}$ and $\Theta_\gamma = \Theta_{ad} \backslash \Theta_\beta$, where $\Theta_\beta$ includes the shift parameters $\beta_l$ at the layers $l\in L$. 

Assume that within $B_t$, there are some samples exhibiting high confidence according to the trained model.  We utilize these predictions as pseudo-labels $\hat{y}$ for the samples with confidence exceeding a threshold $\kappa$, and compute a standard cross-entropy loss:
\begin{equation}
    \mathcal{L}_{pesudo} = \mathbb{E}_{(x)\in \mathcal{D}_{conf}} L_{ce}(\Theta(x), \hat{y}). 
\label{eq:L}
\end{equation}
To circumvent oversimplified solutions by entropy minimization, we propose incorporating an entropy maximization step to update $\Theta_\beta$ on the remaining samples with low confidence: 
\begin{equation}
    \mathcal{L}_{em} = -\mathbb{E}_{(x)\in B_t\backslash\mathcal{D}_{conf}} L_{ent}(\Theta(x))   
\label{eq:H}
\end{equation}
We assume the existence of a single domain-invariant prototype for each class, where all target features should closely align with the source distribution. Intuitively, we encourage a uniform output probability and prevent suboptimal solutions that are overly simplified by maximizing entropy to align target features closer to the coordinate origin. Conversely, to separate distinct features on unlabeled target examples, we propose to minimize entropy on these samples to update $\Theta_\gamma$.
Therefore, our approach can be formulated as a adversarial learning process between $\Theta_\beta$ and $\Theta_\gamma$: $\Theta_\beta$ is trained to maximize the entropy, while $\Theta_\gamma$ is trained to minimize it according to: 
\begin{equation}
\begin{split}
   \Theta_\beta &= arg\min_{\Theta_\beta}\mathcal{L}_{pesudo} - \lambda \mathcal{L}_{em} \\
   \Theta_\gamma &= arg\min_{\Theta_\gamma}\mathcal{L}_{pesudo} + \lambda \mathcal{L}_{em}, 
\end{split}
\label{eq:sup}
\end{equation}
where $\lambda$ is a hyperparameter controlling the trade-off between the minimax entropy and pseudo-labeling objectives. 
%In our implementation, we set $L$ as the final BN layer of the model, and $\lambda=1$. 

\paragraph{Theoretical Insights} Our minimax entropy formulation draws inspiration from Ben-David ~\cite{ben2010theory} and Saito \cite{saito2019semi, saito2018maximum}. Consider a hypothesis $h \in \mathcal{H}$, which can be viewed as a candidate model that approximates a mapping function from inputs to outputs. Let $\epsilon_s(h)$ and $\epsilon_t(h)$ represent the expected risks in the source and target domains, respectively. Then $\epsilon_t(h) \leq \epsilon_s(h) + d_\mathcal{H}(p, q) + \lambda$, where $\lambda$ is a combined error expected to be relatively small, $p$ and $q$ are the source and target distributions, and $d_\mathcal{H}(p, q)$ is the $\mathcal{H}$-divergence between the two distributions, given by:
\begin{equation}
d_\mathcal{H}(p, q) = 2 \sup_{h \in \mathcal{H}} \left| \mathbb{E}_{x_s \sim p} [h(f_s) = 1] - \mathbb{E}_{x_t \sim q} [h(f_t) = 1] \right|.
\label{eq:Hdiv}
\end{equation}
where $f_s$ and $f_t$ denote the features in the source and target domains, respectively. Let $h = \mathds{1}(L_{ent}(\Theta_C(f))\geq\tau)$ be a binary head where $\tau$ is a threshold value and $\Theta_C$ is the classifier. We can rewrite Eq.~\ref{eq:Hdiv} as:
\begin{equation*}
\begin{split}
    d_\mathcal{H}(p, q) &\triangleq 2 \sup_{h\in \mathcal{H}} \left| \mathbf{P_r}_{f_s \sim p}\left[ h(f_s) = 1\right] - \mathbf{P_r}_{f_t \sim q}\left[ h(f_s) = 1\right] \right| \\
    &=  2 | \mathbf{P_r}_{f_s \sim p}\left[ L_{ent}(\Theta_C(f_s)) \geq \tau \right] -  \\
    &\mathbf{P_r}_{f_t \sim q}\left[ L_{ent}(\Theta_C(f_t)) \geq \tau \right] | \\
    &\leq 2 \mathbf{P_r}_{f_t \sim q}\left[ L_{ent}(\Theta_C(f_t)) \geq \tau \right].    
\end{split}
\end{equation*}
We assume $ \mathbf{P_r}_{f_s \sim p}\left[ L_{ent}(\Theta_C(f_s))\geq \tau \right] \leq \mathbf{P_r}_{f_t \sim q}\left[ L_{ent}(\Theta_C(f_t)) \geq \tau \right]$, given that the entropy on a source sample is very small after minimizing it on the main task with $D_s$. Hence, we establish an upper bound by identifying the classifier $\Theta_C$ that achieves maximum entropy for all target features. Our goal is to find the features that achieve the lowest divergence, and the objective can be rewritten as follows:
\[
\min_{f_t} \max_{\Theta_C} \mathbf{P}_{f_t\sim q}\left[ L_{ent}(\Theta_C(f_t)) \geq \tau \right]
\]
Finding the minimum with respect to $f_t$ is equivalent to identifying a feature extractor $\Theta_F$ that achieves this minimum. By setting $\Theta_C = \Theta_\beta$, the maximum entropy optimization on the shift transformation parameters $\{\beta_{l\in L}\}$ serves as a gauge for measuring domain divergence; on the other hand, by setting $\Theta_F = \Theta_\gamma$, the entropy minimization process on the other BN parameters including the scaling parameters $\{\gamma_{l\in L}\}$ and the interpolation weight aims to reduce this divergence. Consequently, we derive the minimax objective of our proposed learning method, as shown in Eq.~\ref{eq:sup}.

\subsubsection{Meta-learning framework} 
\label{sec:meta-l}
In the TTT setting, we assume that we have access to the labeled source domain $D_S=\{D_s\}_{i=1}^M$.  Recent studies \cite{liu2021ttt++, jang2022nearest, Chen2023TTA} indicate that the advantage of TTT is only evident when the descent of SSL loss closely aligns with that of the main loss. To establish this alignment, we leverage a meta-learning paradigm to synchronize gradient descents in both the SSL task and the main task on source data. Each meta-learning iteration involves two steps: meta-train and meta-test. During the meta-train stage, the model focuses on rapid acquisition of domain-specific knowledge through the self-supervised minimax entropy loss. Conversely, in the meta-test stage, the model gradually refines its parameters and the interpolation weight $\alpha$ on the fully supervised main task. This two-step approach allows both tasks to converge and can effectively prevent overfitting of the SSL task. Our full algorithm is outlined in Algorithm~\ref{alg:meta}.
\input{algorithm/meta-learning_1}
\paragraph{Meta-Train\label{par:meta-train}}
$\Theta_{ad}$ is optimized by the proposed SSL loss on $D_s$. When adapting to a new minibatch, we update $\Theta_{ad}$ by alternatively minimizing $\mathcal{L}_{pseudo} - \lambda \mathcal{L}_{em}$ and $\mathcal{L}_{pseudo} + \lambda \mathcal{L}_{em}$, as outlined in Eq.~\ref{eq:sup}. More precisely, we use the former loss to optimize $\Theta_\beta$ and the latter to optimize $\Theta_\gamma$ .
\paragraph{Meta-Test\label{pa:meta-test}}
The meta-test step evaluates the main task with the updated model from meta-train. We use the cross-entropy loss calculated on labeled source data as below:
\begin{equation}
    \mathcal{G} = \mathbb{E}_{(x, y)\in B_t} \mathcal{L}_{ce}(\Theta(x), y),
\label{eq:meta-test}
\end{equation}
where $\Theta = \{\Theta_{ad}, \Theta_{fz}\}$ from meta-train. We update $\Theta_{ad}$ and the interpolation weight $\alpha$ by minimizing $\mathcal{G}$. This necessitates the computation of the second derivative with respect to $\Theta$.

\paragraph{Final-Test} After the model is optimized to convergence on $D_s$. We deploy the final model and conduct the self-supervised "meta-train" step on unlabeled test minibatches. 

\subsection{Batch Domain Shift Synthesis}
\label{sec:method_aug}
To have diverse domains seen at the training stage, we propose a stochastic method for synthesizing domain shifts in a batch. Let $z \in \mathbb{R}^{n \times c \times u \times v}$ represent the input features, where $n$ is the batch size and $c$ is the number of feature channels. To simulate a channel-wise distribution shift, we introduce a linear transformation layer: $\hat{z} = z \cdot \Gamma + \Lambda$, where $\Gamma, \Lambda \in \mathbb{R}^{1 \times c \times 1 \times 1}$ are the weights and biases. Consequently, the synthesized distribution shifts can be formulated as:
\begin{equation}
\begin{split}
    & M \sim \text{Bernoulli}(p), \\
    \Gamma = & (1-M) \odot \Gamma_0 + M \odot R \sim U[0, 1], \\
    \Lambda = & (1-M) \odot \Lambda_0 + M \odot R \sim U[0, 1].
\end{split}
\label{eq:domain_shift_aug}
\end{equation}
where $\odot$ denotes element-wise multiplication. $\Gamma_0$ is set as $1$ and $\Lambda_0$ as $0$. $R \sim U(0,1) \in \mathbb{R}^{1 \times c \times 1 \times 1}$ ranges $[0,1]$. The parameter $p$ denotes a small probability for shifts, while $M$ represents a mask tensor of the same shape as the transformation parameters, i.e., $[1, c, 1, 1]$. This mask tensor determines which channel will have a random distributional statistic shift. In our implementation, we insert the proposed linear transformation layer to induce channel-wise statistical shifts after the stem block within ResNet18 and ResNet50 architectures.

%% file: algorithm/meta-learning_1.tex
\begin{algorithm}[tb]
\caption{Meta-learning for \textit{mixed-BN} Test-Time Training \label{alg:meta}}

\textbf{Input:} Source domain $D_s=\{B_s\}=\{ (x_i, y_i )\}_{i=1}^N$ \\
\textbf{Init:} Model parameters $\Theta$
\begin{algorithmic}[1]
\While {not done}
\State Fetch a minibatch ${B_s}$ from a synthesized domain $D_s \in D_S$ created by Equation~\ref{eq:domain_shift_aug}      
\State Compute $\mathcal{L}_{pesudo}$ and $\mathcal{L}_{em}$ on $B_s$
\State $\Theta_\beta \leftarrow arg\min_{\Theta_\beta}\mathcal{L}_{pesudo} - \lambda \mathcal{L}_{em}$ 
\State $\Theta_\gamma \leftarrow arg\min_{\Theta_\gamma}\mathcal{L}_{pesudo} + \lambda \mathcal{L}_{em}$, 
\State Compute cross entropy loss $\mathcal{G}$ on $B_s$
\State $\Theta_{ad} \leftarrow arg\min_{\Theta_{ad}}\mathcal{G}$
\EndWhile
%\State Update centroids in $q_*$ according to Equation~\ref{eqn: pretrained_centroid}
\State \Return New model parameters $\Theta$ trained on source
\end{algorithmic}
\end{algorithm}

%% file: sec/4_experiments.tex
\section{Experiments}

In this section, we showcase the effectiveness of our method compared to state-of-the-art TTA and TTT techniques on multi-source generalization and image corruption datasets. We also conduct an ablation study to assess the contribution of each component of our approach. Furthermore, we analyze Meta-TTT's performance across different batch sizes and choices of adaptable shift parameters.

\subsection{Experimental Setup}
We compare Meta-TTT to several TTA methods: (1) AdaBN~\cite{li2018AdaBN}, (2) TENT~\cite{wang2021tent}, (3) GEM~\cite{zhang2023adapt}, and (4) ARM~\cite{zhang2021ARM}, all of which are BN-based TTA methods. Additionally, (5) LAME~\cite{boudiaf2022LAME} directly updates output assignment vectors, while (6) TTAC~\cite{su2022revisiting}  updates all network parameters by matching the statistics of the target clusters to the source ones. We also benchmark Meta-TTT against the baseline ERM method.

As for TTT, we select (1) ITTA~\cite{chen2023improved}, (2) TTT-R~\cite{sun2020test}, and (3) TTT++~\cite{liu2021ttt++} for a comparison study. ITTA employs a learnable consistency loss, TTT-R incorporates a rotation head and minimizes loss of rotation prediction, and TTT++ adapts the encoder through online feature alignment combined with a SSL task. 

In our experiments, we use pre-trained ResNet18 and ResNet50 as the model backbone and employ a generalized entropy-minimization loss as in DomainAdaptor \cite{zhang2023adapt}. We designate the final BN layer as the adaptable layer $L$ for $\Theta_C$, set the number of meta-learning steps $k=1$ due to memory constraints, and use a batch size $N=64$.

\input{tables/DG}

\subsection{Comparison to Previous Methods}
\subsubsection{Multi-Source Domain Generalization}
We test Meta-TTT on PACS~\cite{li2017PACS}, OfficeHome~\cite{venkateswara2017OfficeHome}, VLCS~\cite{torralba2011VLCS} and Terra Incognita~\cite{Beery_2018_ECCV} to demonstrate its multi-source generalization performance. PACS consists of 9,991 images from 7 classes across 4 domains (Photo, Art Painting, Cartoon, and Sketch), VLCS contains 10,729 images from 5 classes across 4 domains (Caltech101, LabelMe, SUN09, and VOC2007), OfficeHome has 15,500 images from 65 classes across 4 domains (Art, Clipart, Product, and Real World), and Terra Incognita features 24,788 images spanning 10 classes, collected from 4 different locations, aka. 4 domains (L100, L38, L43, L46). 

We report the results as presented in the respective papers when available; otherwise, we conduct all comparison experiments under the same conditions using the DomainBed~\cite{DomainBed} framework. Further details on hyperparameter selection can be found in the supplementary materials. Based on the ERM results, OfficeHome and Terra Incognita exhibit a large domain gap, while PACS and VLCS have a relatively small domain gap. Table~\ref{tab:DG} shows the comparison to previous methods using Resnet18 as the model backbone. As shown, Meta-TTT achieves state-of-the-art results across all datasets, demonstrating its superior performance in generalizing to unseen domains in the multi-source setting.

\input{tables/ablation_study}
\input{tables/corruption}

\subsubsection{Domain Adaptation to Corruption Datasets}
To understand Meta-TTT's robustness to corruptions, we evaluate it on CIFAR10-C/CIFAR100-C~\cite{hendrycks2019CIFAR10C}, each comprised of 10/100 classes with a training set of 50,000 and a test set of 10,000 examples in 15 corruption types.
% At a larger scale, we evaluate Meta-TTT on ImageNet-C~\cite{hendrycks2019CIFAR10C}, which consists of a training set of 1.2 million in 1,000 classes and 15 types of corruptions on 50,000 test samples.
%~\ref{tab:sp:my-table}

Table~\ref{tab:corruption_cifar10c} and ~\ref{tab:corruption_cifar100c} present the error rates for the highest level of corruption, level 5, on the CIFAR10-C/CIFAR100-C datasets. Except for impulse noise and jpeg compression, Meta-TTT exhibits superior performance across various types of corruption. We attribute our improvements over prior methods to the synchronized descent of the SSL task and the fully supervised main task. In this gradient-based meta-learning paradigm, simultaneous optimization of the SSL loss and the main loss prevents the model from overfitting on the SSL task and fosters robust generalization across diverse domains.

\footnotetext{$D_s$ refers to the entire source data and $B_t$ refers to the target data, which is processed in minibatches sequentially in a streaming manner exactly once. Methods that involve training on both $D_s$ and $B_t$ first train a source model on the modified training objective with $D_s$ and then adapt it on $B_t$ via a SSL task at test time in one pass. Therefore, these methods fall into the Y-O test-time-training protocol according to TTAC's~\cite{su2022revisiting} categorization.} 

\subsection{Ablation Study}
\label{subsec:ablation}
We conduct an ablation study of our proposed Meta-TTT on the PACS dataset to investigate the contributions of individual components, including \textit{mixed-BN}, meta-learning, domain shift augmentation and minimax entropy. We use ResNet18 as the backbone model with a batch size of 64. Each component is added incrementally, and we make the following observations from Table~\ref{tab:ablation_component_effect}. 

\paragraph{\textit{Mixed-BN}:} In ablating \textit{mixed-BN}, we update the affine parameters $\beta$ and $\gamma$ online using entropy minimization while maintaining a fixed interpolation factor $\alpha=0.75$ for Eq. ~\ref{eq:BN}. Applying \textit{Mixed-BN} test-time training to the baseline significantly enhances source model performance from $79.44\%$ (PACS-Source) to $84.02\%$ (PACS-TTT), demonstrating the effectiveness of estimating batch statistics by combining information from both source and target domains.

\paragraph{Meta-L:} When the gradient-based meta-learning paradigm is employed to simultaneously learn from the main task and the self-supervised entropy minimization task, performance improves for both the trained source model PACS-Source (from $79.44\%$ to $83.96\%$) and its test-time training version PACS-TTT (from $84.02\%$ to $85.19\%$). This demonstrates that leveraging meta-learning to align the self-supervised and main tasks not only enhances \textit{mixed-BN} test-time training on test batches but also improves the model's generalization ability to unseen domains.

\paragraph{Shift-Aug:} We utilize a stochastic method to augment domain shift variability during our meta-train process as in Section~\nameref{sec:method_aug}. This augmentation yields an enhancement in both PACS-Source (from $83.96\%$ to $85.13\%$) and PACS-TTT (from $85.19\%$ to $85.39\%$), though the effect is more marginal for PACS-TTT.

\paragraph{Minimax:} Replacing entropy minimization with our minimax entropy approach as outlined in Section~\nameref{sec:method_minimax} yields notable performance enhancements from $85.13\%$ to $85.44\%$ on PACS-Source and from $85.39\%$ to $86.69\%$ on PACS-TTT. We treat $\alpha$ as a learnable interpolation weight instead of a fixed hyperparamter during the meta-test phase on the entire source domains. Compared to the traditional ERM, optimizing our proposed minimax entropy continually enhances model performance over several iterations and effectively prevents oversimplified solutions, resulting in better overall performance. Our experiments employ the complete version of the proposed method, integrating \textit{mixed-BN}, meta-learning, domain shift augmentation and minimax entropy.

\input{tables/batchsize}

\subsection{Additional Analysis}
\paragraph{Robustness to Batch Size}
In table~\ref{tab:batchsize_performance}, we compare prior TTA and TTT methods with Meta-TTT to demonstrate our robustness across various test batch sizes ranging from 16 to 256. When limited by batch sizes, previous works fail to estimate correct batch statistics, possibly due to model overfitting on small test distribution. In contrast, Meta-TTT utilizes adaptable BN parameters, thus performing exceptionally well even for smaller batch sizes. 

\input{tables/bn_layers}
\paragraph{Selection of $\Theta_\beta$}
In our implementation of the minimax strategy in Section~\nameref{sec:method_minimax}, two disjoint set of adaptable parameters $\Theta_\beta$ and $\Theta_\gamma$ are learned adversarially whereby $\Theta_\beta$ is trained to maximize the entropy and $\Theta_\gamma$ strives to minimize it. $\Theta_\beta$ comprises shift parameters $\beta_l$ at the BN layers $l\in L$. In selecting $L$, we aim to ensure the effectiveness of $\Theta_\beta$ and minimize the number of learnable variables to prevent overfitting. Table~\ref{tab:bn_layers} gives the results incurred by selecting different layers for $L$. Based on the overall performance and parameter efficiency, we decide to maximize only the last BN layer in all our experiments.

%% file: tables/DG.tex
\begin{table*}[htbp]
\centering
\begin{tabular}{l*{10}{c}}
\toprule[1pt]\midrule[0.3pt]
\textbf{Method}     & \textbf{Training inputs \protect\footnotemark}  & \textbf{PACS}  & \textbf{VLCS}   & \textbf{OH}   & \textbf{TerraInc}  & \textbf{Avg.} \\ \midrule 
ERM                           & $D_s$     & $79.44_{\pm0.44}$ & $75.77_{\pm0.29}$ & $64.61_{\pm0.18}$ & $39.25_{\pm0.58}$ & 64.77 \\  
AdaBN \cite{li2018AdaBN}       & $B_t$     & $80.44_{\pm0.29}$ & $69.44_{\pm0.48}$ & $63.38_{\pm0.12}$ &  $38.20_{\pm0.19}$ & 62.87 \\ 
Tent \cite{wang2021tent}      & $B_t$     & $83.56_{\pm0.47}$ & $73.37_{\pm0.31}$ & $64.54_{\pm0.06}$ &  $39.87_{\pm0.60}$ & 65.34 \\   
GEM-T \cite{zhang2023adapt}   & $B_t$     & $85.04_{\pm0.23}$ & $77.54_{\pm0.14}$ & $65.39_{\pm0.19}$ & $42.37_{\pm0.96}$ & 67.59 \\
GEM-SKD \cite{zhang2023adapt} & $B_t$     & $84.37_{\pm0.28}$ & $78.10_{\pm0.14}$ & $65.61_{\pm0.14}$ & $42.45_{\pm1.01}$ & 67.63 \\
GEM-Aug \cite{zhang2023adapt} & $B_t$     & $84.93_{\pm0.19}$ & $78.50_{\pm0.22}$ & $66.73_{\pm0.25}$ & $42.98_{\pm1.19}$ & 68.29 \\
LAME \cite{boudiaf2022LAME}   & $B_t$     & $80.28_{\pm0.33}$ & $75.59_{\pm0.96}$ & $63.16_{\pm0.28}$ &  $38.37_{\pm1.12}$ & 64.35 \\ 
ARM \cite{zhang2021ARM}       & $D_s+B_t$ & $82.47_{\pm0.59}$ & $68.21_{\pm1.68}$ & $63.15_{\pm0.61}$ & $37.60_{\pm2.39}$ & 62.86 \\ 
TTAC \cite{su2022revisiting}  & $B_t$ & $82.35_{\pm0.50}$ & $74.74_{\pm1.08}$ & $63.13_{\pm0.52}$ & $42.19_{\pm1.69}$ & 65.60 \\
% CoTTA \cite{wang2022COTTA}    & $B_t$     &        &        &        &      \\
\midrule
ITTA  \cite{chen2023improved}      & $D_s+B_t$ & $82.25_{\pm0.54}$ & $72.28_{\pm2.48}$ & $66.42_{\pm0.48}$ & $41.07_{\pm1.14}$ & 65.51 \\
TTT-R \cite{sun2020test}      & $D_s+B_t$ & $83.35_{\pm0.57}$ & $77.15_{\pm0.35}$ & $66.37_{\pm0.28}$ & $35.37_{\pm0.74}$ & 65.56 \\
TTT++ \cite{liu2021ttt++}     & $D_s+B_t$ & $82.11_{\pm0.75}$ & $73.84_{\pm0.16}$ & $61.84_{\pm0.24}$ & $43.52_{\pm0.31}$ & 65.33 \\ \midrule 
Ours                      & $D_s+B_t$ & $\textbf{86.69}_{\pm\textbf{0.03}}$ & $\textbf{78.73}_{\pm\textbf{0.41}}$ & $\textbf{67.36}_{\pm\textbf{0.45}}$ & $\textbf{43.77}_{\pm\textbf{0.27}}$ & \textbf{69.14} \\ 
\midrule[0.3pt]\bottomrule[1pt]
\end{tabular}
\caption{Comparison of multi-source generalization results using ResNet18 as the backbone model. \label{tab:DG}}
\vspace{-3mm}
\end{table*} 

%% file: tables/ablation_study.tex
\begin{table}[ht]
\centering
\begin{adjustbox}{max width=\linewidth}
\begin{tabular}{c c c c | c c }
\toprule[1pt]\midrule[0.3pt]
\textit{Mixed-BN} & Meta-L & Shift-Aug  & Minimax & PACS-TTT & PACS-Source \\ \midrule 
\checkmark & & & & 84.02 & 79.44 \\
\checkmark & \checkmark & & & 85.19 & 83.96 \\
\checkmark & \checkmark & \checkmark & & 85.39 & 85.13\\
\checkmark & \checkmark & \checkmark &  \checkmark & \textbf{86.69} & \textbf{85.44} \\
\midrule[0.3pt]\bottomrule[1pt]
\end{tabular}
\end{adjustbox}
\caption{Ablation study of our method with ResNet18. PACS-TTT and PACS-Source represent the model performance on target domain with and without the adaptation at test time, respectively. \label{tab:ablation_component_effect}}
\vspace{-3mm}
\end{table}

%% file: tables/corruption.tex
\begin{table*}[htb]
\centering
\begin{adjustbox}{max width=1\textwidth}{%
\begin{tabular}{l*{17}{c}}
\toprule[1pt]\midrule[0.3pt]
\textbf{} & \textbf{Avg. err} & \textbf{Gaus.} & \textbf{Shot} & \textbf{Impu.} & \textbf{Defo.} & \textbf{Glas.} & \textbf{Moti.} & \textbf{Zoom} & \textbf{Snow} & \textbf{Fros.} & \textbf{Fog} & \textbf{Brig.} & \textbf{Cont.} & \textbf{Elas.} & \textbf{Pixe.} & \textbf{Jpeg} \\  \midrule
\textbf{Source} & 36.63 & 44.19 & 41.16 & 53.02 & 32.65 & 57.59 & 32.83 & 26.85 & 30.25 & 34.10 & 31.20 & 14.72 & 34.04 & 42.80 & 36.79 & 37.19 \\
\textbf{Tent} \cite{wang2021tent} & 30.99 & 38.01 & 34.27 & 47.89 & 23.44 & 57.57 & 26.50 & 19.18 & 24.88 & 29.11 & 25.08 & 11.31 & 27.05 & 37.81 & 31.90 & 30.91 \\
\textbf{TTAC} \cite{su2022revisiting} & 18.53 & 27.12 & 24.39 & \textbf{21.78} & 17.31 & 28.70 & 15.72 & 13.07 & 16.52 & 18.51 & 17.75 & 7.55 & 10.62 & 25.05 & 14.56 & \textbf{19.30} \\ 
\textbf{TTT-R} \cite{sun2020test} & 28.72 & 35.39 & 32.87 & 41.70 & 21.03 & 42.16 & 27.06 & 17.69 & 24.48 & 22.07 & 25.68 & 14.08 & 33.78 & 32.08 & 37.85 & 22.95 \\
\textbf{TTT++} \cite{liu2021ttt++} & 36.41 & 45.89 & 42.18 & 53.32 & 29.78 & 57.73 & 31.39 & 25.87 & 30.25 & 34.99 & 32.03 & 14.22 & 33.58 & 42.90 & 36.17 & 35.81 \\ \midrule 
\textbf{Ours} & \textbf{14.87} & \textbf{25.15} & \textbf{23.46} & 36.60 & \textbf{8.02} & \textbf{21.49} & \textbf{10.65} & \textbf{5.93} & \textbf{11.20} & \textbf{10.29} &\textbf{ 10.94} & \textbf{5.36} & \textbf{7.89} & \textbf{15.21} & \textbf{8.38} & 22.52 \\ 
\midrule[0.3pt]\bottomrule[1pt]
\end{tabular}}
\end{adjustbox}
\caption{Error rates across various corruptions on the severity level 5 of CIFAR10-C with Resnet50 \label{tab:corruption_cifar10c}}
\end{table*}

\begin{table*}[htb]
\centering
\begin{adjustbox}{max width=1\textwidth}
\begin{tabular}{l*{17}{c}}
\toprule[1pt]\midrule[0.3pt]
\textbf{} & \textbf{Avg. err} & \textbf{Gaus.} & \textbf{Shot} & \textbf{Impu.} & \textbf{Defo.} & \textbf{Glas.} & \textbf{Moti.} & \textbf{Zoom} & \textbf{Snow} & \textbf{Fros.} & \textbf{Fog} & \textbf{Brig.} & \textbf{Cont.} & \textbf{Elas.} & \textbf{Pixe.} & \textbf{Jpeg} \\ \midrule
\textbf{Source} & 61.94 & 71.54 & 69.73 & 83.00 & 53.91 & 78.57 & 54.58 & 49.83 & 59.35 & 58.93 & 59.88 & 39.84 & 52.98 & 67.91 & 63.14 & 65.94 \\
\textbf{Tent} \cite{wang2021tent} & 54.33 & 71.09 & 61.95 & 92.04 & 37.30 & 85.42 & 39.03 & 33.76 & 58.38 & 51.61 & 44.34 & 27.60 & 39.34 & 68.65 & 52.95 & 51.56 \\
\textbf{TTAC} \cite{su2022revisiting} & 48.05 & 56.24 & 55.32 & \textbf{60.33} & 40.69 & 62.51 & 42.98 & 38.69 & 49.11 & 50.27 & 46.51 & 33.97 & 35.27 & 52.54 & 48.55 & \textbf{47.79} \\
\textbf{TTT-R} \cite{sun2020test} & 48.95 & 63.98 & 62.15 &  67.76 & 36.31 & 63.02 & 42.92 & 31.78 & 43.19 & 42.10 & 45.82 & 30.18 & 52.57 & 50.33 & 52.75 & 49.44 \\
\textbf{TTT++} \cite{liu2021ttt++} & 61.45 & 72.88 & 71.54 & 83.73 & 52.26 & 78.69 & 54.55 & 49.04 & 59.85 & 60.84 & 44.34 & 40.76 & 54.47 & 68.03 & 63.98 & 66.76 \\  \midrule
\textbf{Ours} & \textbf{38.46} & \textbf{54.11} &  \textbf{52.89} & 64.50 & \textbf{26.99} & \textbf{48.10} & \textbf{31.65} & \textbf{23.97} & \textbf{35.19} & \textbf{33.92} & \textbf{35.96} & \textbf{22.93} & \textbf{28.88} & \textbf{39.00} & \textbf{29.10} & 49.72\\ 
\midrule[0.3pt]\bottomrule[1pt]
\end{tabular}
\end{adjustbox}
\caption{Error rates across various corruptions on the severity level 5 of CIFAR100-C with Resnet50 \label{tab:corruption_cifar100c}}
\end{table*}

%% file: tables/batchsize.tex
\begin{table}[H]
\centering
\begin{adjustbox}{max width=\linewidth}
\begin{tabular}{l*{6}{c}}
\toprule[1pt]\midrule[0.3pt]
\textbf{} & \textbf{16} & \textbf{32} & \textbf{64} & \textbf{256} \\ \midrule
\textbf{Tent} \cite{wang2021tent} & 38.04 & 38.02 & 38.01 & 37.97 \\ 
\textbf{GEM-T} \cite{zhang2023adapt} & 35.27 & 35.15 & 35.13 &  34.90 \\ 
\textbf{TTT-R} \cite{sun2020test} & 36.21 & 35.48 & 35.39 & 35.36 \\ 
\textbf{TTT++} \cite{liu2021ttt++} & 46.55 & 46.02 & 45.89 &  45.86 \\ \midrule
\textbf{Ours} & 25.29 & 25.15 & 25.15 & 25.13 \\
\midrule[0.3pt]\bottomrule[1pt]
\end{tabular}
\end{adjustbox}
\caption{Error rates for different test batch sizes (16, 32, 64 and 256) on CIFAR10-C with ResNet50 under Gaussian noise corruption (severity level 5) \label{tab:batchsize_performance}}
\end{table}

%% file: tables/bn_layers.tex
\begin{table}[ht]
\centering
\begin{tabular}{ c c | c }
\toprule[1pt]\midrule[0.3pt]
BN Layers & Params & Error rate (\%) \\ \midrule 
None & 0 & 27.77 \\
Last & 2,048 & \textbf{25.15} \\
Layer4 & 9,216 & 25.91 \\
BN3 & 15,104 & 25.67 \\
All & 22,720 & 26.70 \\
\midrule[0.3pt]\bottomrule[1pt]
\end{tabular}
\caption{Selection of BN layers to maximize the entropy based on error rate at the most severe level of the Gaussian corruption type of CIFAR10-C (using a ResNet50). \textit{None} means using the traditional ERM method for all BNs; \textit{Last} refers to the very last BN layer; \textit{Layer4} refers to BN layers in the last block of the ResNet50 backbone; \textit{BN3} refers to the third BN layers. \textit{All} means applying entropy maximization at all BN layers. \label{tab:bn_layers}}
\vspace{-3mm}
\end{table}

%% file: sec/5_conclusion.tex
\section{Conclusion}
In this paper, we present a novel meta-learning framework for \textit{mixed-BN} based test-time training. Our method aims to improve existing TTT strategies in three aspects: a novel meta-learning framework, minimax entropy, and domain shift augmentation. Through extensive experiments, we show that our method can achieve superior performance on both the multi-source domain generalization and single-source domain adaptation tasks. Future work will focus on continual adaptation at test time \cite{wang2022COTTA} to cater to real-world scenarios where the test distribution can come from continually changing domains and are not drawn independently and identically (non-i.i.d.) \cite{yuan2023robust, gong2022note}.

%% file: sec/X_suppl.tex
\clearpage
\setcounter{page}{1}
\appendix
\section{Implementation Details}

We choose soft cross-entropy ($L_{ce}$ in Equation~\ref{eq:L}) for the pseudo labeling loss and the Generalized Entropy Minimization (GEM-T) \cite{zhang2023adapt} loss in Equation~\ref{eq:H} to optimize the SSL task. Note that although GEM-Aug performs the best among the GEM family according to DomainAdaptor \cite{zhang2023adapt}, we avoid using it since the augmentation loss counteracts and cancels out the effects of our proposed domain shift augmentation. We utilize SGD optimizers with Nesterov accelerated gradient \cite{Nesterov1983AMF} for both the meta-train and meta-test phases. Additional hyperparameters are set as follows: learning rate = $0.001$, meta learning rate = $0.05$, learning rate decay = $0.1$, learning rate for the final classifier network = $0.01$, momentum = $0.9$, weight decay = $0.0005$, initial value for the interpolation weight = $0.75$ and its learning rate = $0.1$. For image transformation, we apply color jittering, resize the image dimensions to 224 x 224 and perform a random resized cropping with a scale of $0.8$.

For multi-source domain generalization datasets (PACS, VLCS, OH and TerraInc) in Table~\ref{tab:DG}, we follow the leave-one-domain-out cross-validation method in DomainBed~\cite{DomainBed} for hyperparameter selection. Specifically, given \textit{n} training domains, we leave out one domain at a time and evaluate the model trained on the remaining $n-1$ domains on that held-out domain. We then average the accuracies across all held-out domains. The set of hyperparameters that maximizes this average accuracy is selected, and we retrain the model using these hyperparameters in the same leave-one-domain-out manner. We average the results from three independent runs with different random seeds, and maintain a fixed batch size of 64. Table~\ref{tab:hyperparams} describes our hyperparameter search spaces over which we conduct 20 random trials. Due to computational limitation and the fact that only a portion of hyperparameters affect model performance to a noticeable degree, we only search through learning rates and the initialization of interpolation weight. The remaining hyperparameters default to the standard values described earlier in this section.

\input{tables/hyperparams}

\section{Further Analysis\label{sec:sp:analysis}}
\subsection{Efficacy of Meta Learning }
\input{tables/MetaLforTTT}
In this study, we propose leveraging the gradient-based meta-learning framework to align the SSL task with the primary task when training the source model. We further explore the efficacy of this framework across various prevalent SSL tasks.

Table~\ref{tab:sp:metalearning_with_SSL} presents TTT results of some of the popular SSL tasks, including rotation identification and norm minimization. The findings indicate that the meta-learning approach effectively aligns multiple SSL tasks with the primary task and leads to performance enhancements of source models.

\subsection{Efficacy of Minimax Entropy}
\begin{figure}[!ht]
  \includegraphics[scale=0.7]{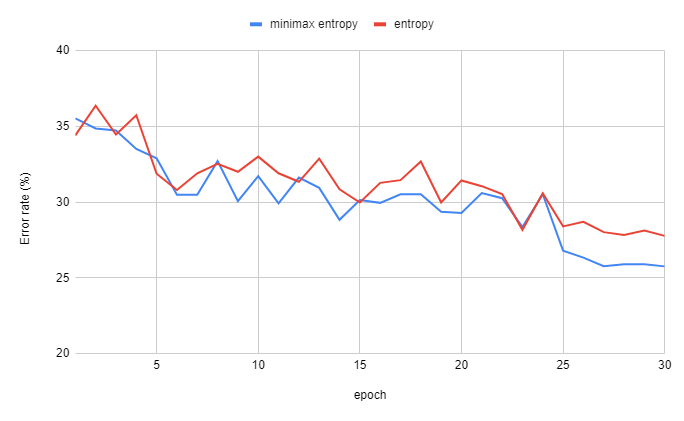}
  \caption{Training curves comparing minimax entropy and traditional entropy (ERM) on Gaussian noise corruption at the highest severity on CIFAR10-C}
  \label{fig:minimax}
\end{figure}

Figure~\ref{fig:minimax} demonstrates the superiority of minimax entropy over the standard entropy. At the end of each meta-train epoch, we adapt the model checkpoint trained on CIFAR10 to the most severe Gaussian corruption type in CIFAR10-C. The trend indicates that the model optimized using minimax entropy consistently improves its performance on unseen corrupted data, proven to be more effective than the traditional entropy method.

%% file: tables/hyperparams.tex
\begin{table}[h]
\centering
\begin{adjustbox}{max width=\linewidth}
\begin{tabular}{ c c c }
\toprule[1pt]\midrule[0.3pt]
Parameter & Best Value & Search Space \\ \midrule 
initialization of $\alpha$ & 0.75 & $[0.55, 0.65, 0.75, 0.85]$ \\
learning rate for $\alpha$ & 0.1 & $[0.005, 0.1, 0.5]$ \\
meta-learning rate & 0.05 & $[0.01, 0.05, 0.1]$ \\
learning rate (others) & 0.001 & $[0.0005, 0.001, 0.005, 0.01]$ \\
\midrule[0.3pt]\bottomrule[1pt]
\end{tabular}
\end{adjustbox}
\caption{Hyperparameter search spaces and their best values for our method. We maintain a learning rate for the final classifier network to be ten times the original learning rate. \label{tab:hyperparams}}
\vspace{-3mm}
\end{table}

%% file: tables/MetaLforTTT.tex
\begin{table}
\centering
\setlength{\tabcolsep}{2pt}
\begin{adjustbox}{max width=1\linewidth}
\begin{tabular}{l*{10}{c}}
\toprule[1pt]\midrule[0.3pt]
\cmidrule(lr){7-11}  
& $\mathbf{Art}$ & $\mathbf{Cartoon}$ & $\mathbf{Photo}$ & $\mathbf{sketch}$ &\textbf{Avg}
\\ \midrule
TTT-Rotation  &  78.66 &  80.80 &  95.81 & 78.14 & 83.35 \\
% & 86.77 & 83.40 &98.20&81.37&87.44\\
Meta-L + TTT-Rotation  & \textbf{82.71} & \textbf{81.44} & \underline{96.65} & \underline{78.72} & \textbf{84.88} \\ \midrule
TTT-Norm    & 80.22 & 80.33 & 96.35 & 77.60 & 83.63 \\
Meta-L + TTT-Norm & \underline{82.62} & \underline{80.16} & \textbf{96.77} & \textbf{78.34} & \underline{84.47}  \\  
%MetaL+MME       & 85.01 & 80.57 & 97.02 & 81.51 & 86.03 \\
\midrule[0.3pt]\bottomrule[1pt]
\end{tabular}
\end{adjustbox}
\caption{Efficacy of the meta-learning framework with different self-supervised objectives for TTT on PACS with ResNet18.\label{tab:sp:metalearning_with_SSL}}
\vspace{-4mm}
\end{table}

%% file: main.bbl
\begin{thebibliography}{54}
\providecommand{\natexlab}[1]{#1}

\bibitem[{Balaji, Sankaranarayanan, and Chellappa(2018)}]{balaji2018metareg}
Balaji, Y.; Sankaranarayanan, S.; and Chellappa, R. 2018.
\newblock Metareg: Towards domain generalization using metaregularization.
\newblock In \emph{NeurIPS}.

\bibitem[{Beery, Van~Horn, and Perona(2018)}]{Beery_2018_ECCV}
Beery, S.; Van~Horn, G.; and Perona, P. 2018.
\newblock Recognition in Terra Incognita.
\newblock In \emph{Proceedings of the European Conference on Computer Vision (ECCV)}.

\bibitem[{Ben-David et~al.(2010)Ben-David, Blitzer, Crammer, Kulesza, Pereira, and Vaughan}]{ben2010theory}
Ben-David, S.; Blitzer, J.; Crammer, K.; Kulesza, A.; Pereira, F.; and Vaughan, J.~W. 2010.
\newblock A theory of learning from different domains.
\newblock \emph{Machine learning}, 79: 151--175.

\bibitem[{Boudiaf et~al.(2022)Boudiaf, Mueller, Ben~Ayed, and Bertinetto}]{boudiaf2022LAME}
Boudiaf, M.; Mueller, R.; Ben~Ayed, I.; and Bertinetto, L. 2022.
\newblock Parameter-free online test-time adaptation.
\newblock In \emph{CVPR}.

\bibitem[{Chen et~al.(2023{\natexlab{a}})Chen, Zhang, Song, Shan, and Liu}]{Chen2023TTA}
Chen, L.; Zhang, Y.; Song, Y.; Shan, Y.; and Liu, L. 2023{\natexlab{a}}.
\newblock Continual test-time domain adaptation.
\newblock In \emph{CVPR}, 24172--24182.

\bibitem[{Chen et~al.(2023{\natexlab{b}})Chen, Zhang, Song, Shan, and Liu}]{chen2023improved}
Chen, L.; Zhang, Y.; Song, Y.; Shan, Y.; and Liu, L. 2023{\natexlab{b}}.
\newblock Improved Test-Time Adaptation for Domain Generalization.
\newblock In \emph{CVPR}.

\bibitem[{Choi et~al.(2022)Choi, Yang, Choi, and Yun}]{choi2022SWR}
Choi, S.; Yang, S.; Choi, S.; and Yun, S. 2022.
\newblock Improving test-time adaptation via shift-agnostic weight regularization and nearest source prototypes.
\newblock In \emph{ECCV}, 440--458. Springer.

\bibitem[{Deng et~al.(2020)Deng, Ding, Dwork, Hong, Parmigiani, Patil, and Sur}]{deng2020representation}
Deng, Z.; Ding, F.; Dwork, C.; Hong, R.; Parmigiani, G.; Patil, P.; and Sur, P. 2020.
\newblock Representation via representations: Domain generalization via adversarially learned invariant representations.
\newblock \emph{arXiv preprint arXiv:2008.00601}.

\bibitem[{Dou et~al.(2019)Dou, de~Castro, Kamnitsas, and Glocker}]{dou2019domain}
Dou, Q.; de~Castro, D.~C.; Kamnitsas, K.; and Glocker, B. 2019.
\newblock Domain generalization via model-agnostic learning of semantic features.
\newblock In \emph{NeurIPS}.

\bibitem[{Finn, Abbeel, and Levine(2017)}]{finn2017MAML}
Finn, C.; Abbeel, P.; and Levine, S. 2017.
\newblock Model-agnostic meta-learning for fast adaptation of deep networks.
\newblock In \emph{ICML}.

\bibitem[{Fu et~al.(2021)Fu, Liu, Iqbal, Mello, Shi, and Kautz}]{Fu2021LearningTT}
Fu, Y.; Liu, S.; Iqbal, U.; Mello, S.~D.; Shi, H.; and Kautz, J. 2021.
\newblock Learning to Track Instances without Video Annotations.
\newblock \emph{2021 IEEE/CVF Conference on Computer Vision and Pattern Recognition (CVPR)}, 8676--8685.

\bibitem[{Gong et~al.(2022)Gong, Jeong, Kim, Kim, Shin, and Lee}]{gong2022note}
Gong, T.; Jeong, J.; Kim, T.; Kim, Y.; Shin, J.; and Lee, S.-J. 2022.
\newblock NOTE: Robust continual test-time adaptation against temporal correlation.
\newblock In \emph{NeurIPS}.

\bibitem[{Gulrajani and Lopez-Paz(2021{\natexlab{a}})}]{gulrajani2021in}
Gulrajani, I.; and Lopez-Paz, D. 2021{\natexlab{a}}.
\newblock In Search of Lost Domain Generalization.
\newblock In \emph{ICLR}.

\bibitem[{Gulrajani and Lopez-Paz(2021{\natexlab{b}})}]{DomainBed}
Gulrajani, I.; and Lopez-Paz, D. 2021{\natexlab{b}}.
\newblock In search of lost domain generalization.
\newblock In \emph{PICLR}.

\bibitem[{Hansen et~al.(2020)Hansen, Sun, Abbeel, Efros, Pinto, and Wang}]{Hansen2020SelfSupervisedPA}
Hansen, N.; Sun, Y.; Abbeel, P.; Efros, A.~A.; Pinto, L.; and Wang, X. 2020.
\newblock Self-Supervised Policy Adaptation during Deployment.
\newblock \emph{ArXiv}, abs/2007.04309.

\bibitem[{Hendrycks and Dietterich(2019)}]{hendrycks2019CIFAR10C}
Hendrycks, D.; and Dietterich, T. 2019.
\newblock Benchmarking neural network robustness to common corruptions and perturbations.
\newblock \emph{arXiv preprint arXiv:1903.12261}.

\bibitem[{Hoffman et~al.(2018)Hoffman, Tzeng, Park, Zhu, Isola, Saenko, Efros, and Darrell}]{hoffman2018cycada}
Hoffman, J.; Tzeng, E.; Park, T.; Zhu, J.-Y.; Isola, P.; Saenko, K.; Efros, A.~A.; and Darrell, T. 2018.
\newblock CyCADA: Cycle-consistent adversarial domain adaptation.
\newblock In \emph{ICML}, 1989--1998.

\bibitem[{Jang, Chung, and Chung(2022)}]{jang2022nearest}
Jang, M.; Chung, S.-Y.; and Chung, H.~W. 2022.
\newblock Test-Time Adaptation via Self-Training with Nearest Neighbor Information.
\newblock In \emph{ICLR}.

\bibitem[{Kang et~al.(2019)Kang, Jiang, Yang, and Hauptmann}]{kang2019contrastive}
Kang, G.; Jiang, L.; Yang, Y.; and Hauptmann, A.~G. 2019.
\newblock Contrastive adaptation network for unsupervised domain adaptation.
\newblock In \emph{CVPR}.

\bibitem[{Li et~al.(2017)Li, Yang, Song, and Hospedales}]{li2017PACS}
Li, D.; Yang, Y.; Song, Y.-Z.; and Hospedales, T.~M. 2017.
\newblock Deeper, broader and artier domain generalization.
\newblock In \emph{ICCV}.

\bibitem[{Li et~al.(2018{\natexlab{a}})Li, Yang, Song, and Hospedales}]{li2018MLDG}
Li, D.; Yang, Y.; Song, Y.-Z.; and Hospedales, T.~M. 2018{\natexlab{a}}.
\newblock Learning to generalize: Meta-learning for domain generalization.
\newblock In \emph{AAAI}.

\bibitem[{Li et~al.(2018{\natexlab{b}})Li, Pan, Wang, and Kot}]{li2018CVPR}
Li, H.; Pan, S.; Wang, S.; and Kot, A. 2018{\natexlab{b}}.
\newblock Domain generalization with adversarial feature learning.
\newblock In \emph{CVPR}.

\bibitem[{Li et~al.(2021)Li, Li, Li, Gong, Fu, and Hospedales}]{li2021simple}
Li, P.; Li, D.; Li, W.; Gong, S.; Fu, Y.; and Hospedales, T.~M. 2021.
\newblock A simple feature augmentation for domain generalization.
\newblock In \emph{ICCV}.

\bibitem[{Li et~al.(2020)Li, Jiao, Cao, and Wong}]{li2020cvpr}
Li, R.; Jiao, Q.; Cao, W.; and Wong, S., H.S.and~Wu. 2020.
\newblock Model adaptation: Unsuper-vised domain adaptation without source data.
\newblock In \emph{CVPR}.

\bibitem[{Li et~al.(2018{\natexlab{c}})Li, Wang, Shi, Hou, and Liu}]{li2018AdaBN}
Li, Y.; Wang, N.; Shi, J.; Hou, X.; and Liu, J. 2018{\natexlab{c}}.
\newblock Adaptive batch normalization for practical domain adaptation.
\newblock \emph{PR}.

\bibitem[{Liang, Hu, and Feng(2020)}]{liang2020SHOT}
Liang, J.; Hu, D.; and Feng, J. 2020.
\newblock Do we really need to access the source data? source hypothesis transfer for unsupervised domain adaptation.
\newblock In \emph{ICML}, 6028--6039.

\bibitem[{Lim et~al.(2023)Lim, Kim, Choo, and Choi}]{lim2023ttn}
Lim, H.; Kim, B.; Choo, J.; and Choi, S. 2023.
\newblock TTN: A Domain-Shift Aware Batch Normalization in Test-Time Adaptation.
\newblock In \emph{ICLR}.

\bibitem[{Liu et~al.(2021)Liu, Kothari, Van~Delft, Bellot-Gurlet, Mordan, and Alahi}]{liu2021ttt++}
Liu, Y.; Kothari, P.; Van~Delft, B.; Bellot-Gurlet, B.; Mordan, T.; and Alahi, A. 2021.
\newblock Ttt++: When does self-supervised test-time training fail or thrive?
\newblock \emph{Advances in Neural Information Processing Systems}, 34: 21808--21820.

\bibitem[{Luo et~al.(2019)Luo, Zheng, Guan, Yu, and Yang}]{luo2019taking}
Luo, Y.; Zheng, L.; Guan, T.; Yu, J.; and Yang, Y. 2019.
\newblock Taking a closer look at domain shift: Category-level adversaries for semantics consistent domain adaptation.
\newblock In \emph{CVPR}.

\bibitem[{Nam et~al.(2021)Nam, Lee, Park, Yoon, and Yoo}]{nam2021reducing}
Nam, H.; Lee, H.; Park, J.; Yoon, W.; and Yoo, D. 2021.
\newblock Reducing domain gap by reducing style bias.
\newblock In \emph{CVPR}.

\bibitem[{Nesterov(1983)}]{Nesterov1983AMF}
Nesterov, Y. 1983.
\newblock A method for solving the convex programming problem with convergence rate O$(1/k^2)$.
\newblock In \emph{Proceedings of the USSR Academy of Sciences}, volume 269, 543--547.

\bibitem[{Niu et~al.(2023)Niu, Wu, Zhang, Wen, Chen, Zhao, and Tan}]{niu2023}
Niu, S.; Wu, J.; Zhang, Y.; Wen, Z.; Chen, Y.; Zhao, P.; and Tan, M. 2023.
\newblock Towards Stable Test-Time Adaptation in Dynamic Wild World.
\newblock In \emph{ICLR}.

\bibitem[{Rame, Dancette, and Cord(2022)}]{rame2022fishr}
Rame, A.; Dancette, C.; and Cord, M. 2022.
\newblock Fishr: Invariant gradient variances for out-of-distribution generalization.
\newblock In \emph{ICML}.

\bibitem[{Saito et~al.(2019)Saito, Kim, Sclaroff, Darrell, and Saenko}]{saito2019semi}
Saito, K.; Kim, D.; Sclaroff, S.; Darrell, T.; and Saenko, K. 2019.
\newblock Semi-supervised domain adaptation via minimax entropy.
\newblock In \emph{Proceedings of the IEEE/CVF international conference on computer vision}, 8050--8058.

\bibitem[{Saito et~al.(2018)Saito, Watanabe, Ushiku, and Harada}]{saito2018maximum}
Saito, K.; Watanabe, K.; Ushiku, Y.; and Harada, T. 2018.
\newblock Maximum classifier discrepancy for unsupervised domain adaptation.
\newblock In \emph{CVPR}, 3723--3732.

\bibitem[{Shi et~al.(2021)Shi, Seely, Torr, Siddharth, Hannun, Usunier, and Synnaeve}]{shi2021gradient}
Shi, Y.; Seely, J.; Torr, P.~H.; Siddharth, N.; Hannun, A.; Usunier, N.; and Synnaeve, G. 2021.
\newblock Gradient matching for domain generalization.
\newblock In \emph{ICLR}.

\bibitem[{Singh and Shrivastava(2019)}]{singh2019evalnorm}
Singh, S.; and Shrivastava, A. 2019.
\newblock Evalnorm: Estimating batch normalization statistics for evaluation.
\newblock In \emph{ICCV}.

\bibitem[{Su, Xu, and Jia(2022)}]{su2022revisiting}
Su, Y.; Xu, X.; and Jia, K. 2022.
\newblock Revisiting Realistic Test-Time Training: Sequential Inference and Adaptation by Anchored Clustering.
\newblock In Oh, A.~H.; Agarwal, A.; Belgrave, D.; and Cho, K., eds., \emph{Advances in Neural Information Processing Systems}.

\bibitem[{Summers and Dinneen(2019)}]{summers2019four}
Summers, C.; and Dinneen, M.~J. 2019.
\newblock Four things everyone should know to improve batch normalization.
\newblock In \emph{ICLR}.

\bibitem[{Sun, Feng, and Saenko(2016)}]{sun2016Coral}
Sun, B.; Feng, J.; and Saenko, K. 2016.
\newblock Return of frustratingly easy domain adaptation.
\newblock In \emph{AAAI}.

\bibitem[{Sun et~al.(2020)Sun, Wang, Liu, Miller, Efros, and Hardt}]{sun2020test}
Sun, Y.; Wang, X.; Liu, Z.; Miller, J.; Efros, A.; and Hardt, M. 2020.
\newblock Test-time training with self-supervision for generalization under distribution shifts.
\newblock In \emph{ICML}.

\bibitem[{Torralba and Efros(2011)}]{torralba2011VLCS}
Torralba, A.; and Efros, A.~A. 2011.
\newblock Unbiased look at dataset bias.
\newblock In \emph{CVPR}, 1521--1528.

\bibitem[{Venkateswara et~al.(2017)Venkateswara, Eusebio, Chakraborty, and Panchanathan}]{venkateswara2017OfficeHome}
Venkateswara, H.; Eusebio, J.; Chakraborty, S.; and Panchanathan, S. 2017.
\newblock Deep hashing network for unsupervised domain adaptation.
\newblock In \emph{CVPR}, 5018--5027.

\bibitem[{Wang et~al.(2021)Wang, Shelhamer, Liu, Olshausen, and Darrell}]{wang2021tent}
Wang, D.; Shelhamer, E.; Liu, S.; Olshausen, B.; and Darrell, T. 2021.
\newblock Tent: Fully Test-Time Adaptation by Entropy Minimization.
\newblock In \emph{ICLR}.

\bibitem[{Wang et~al.(2022)Wang, Fink, Van~Gool, and Dai}]{wang2022COTTA}
Wang, Q.; Fink, O.; Van~Gool, L.; and Dai, D. 2022.
\newblock Continual test-time domain adaptation.
\newblock In \emph{CVPR}, 7201--7211.

\bibitem[{Yan et~al.(2020)Yan, Song, Li, Zou, and Ren}]{yan2020improve}
Yan, S.; Song, H.; Li, N.; Zou, L.; and Ren, L. 2020.
\newblock Improve unsupervised domain adaptation with mixup training.
\newblock In \emph{arXiv preprint arXiv:2001.00677}.

\bibitem[{Yang et~al.(2021)Yang, Wang, van~de Weijer, J., and L.}]{yang2020iccv}
Yang, S.; Wang, Y.; van~de Weijer; J., H.; and L., S., Jui. 2021.
\newblock Generalized sourcefree domain adaptation.
\newblock In \emph{ICCV}.

\bibitem[{You, Li, and Zhao(2021)}]{you2021alpha}
You, F.; Li, J.; and Zhao, Z. 2021.
\newblock Test-time batch statistics calibration for covariate shift.
\newblock \emph{arXiv preprint arXiv:2110.04065}.

\bibitem[{Yuan, Xie, and Li(2023)}]{yuan2023robust}
Yuan, L.; Xie, B.; and Li, S. 2023.
\newblock Robust test-time adaptation in dynamic scenarios.
\newblock In \emph{CVPR}.

\bibitem[{Zhang et~al.(2023)Zhang, Qi, Shi, and Gao}]{zhang2023adapt}
Zhang, J.; Qi, L.; Shi, Y.; and Gao, Y. 2023.
\newblock DomainAdaptor: A Novel Approach to Test-time Adaptation.
\newblock In \emph{ICCV}, 18971--18981.

\bibitem[{Zhang, Levine, and Finn(2022)}]{zhang2022memo}
Zhang, M.; Levine, S.; and Finn, C. 2022.
\newblock Memo: Test time robustness via adaptation and augmentation.
\newblock In \emph{NeurIPS}.

\bibitem[{Zhang et~al.(2021)Zhang, Marklund, Dhawan, Gupta, Levine, and Finn}]{zhang2021ARM}
Zhang, M.; Marklund, H.; Dhawan, N.; Gupta, A.; Levine, S.; and Finn, C. 2021.
\newblock Adaptive risk minimization: Learning to adapt to domain shift.
\newblock \emph{Advances in Neural Information Processing Systems}, 34: 23664--23678.

\bibitem[{Zhang et~al.(2022)Zhang, Shen, Zhang, and Foo}]{zhang2022lccs}
Zhang, W.; Shen, L.; Zhang, W.; and Foo, C.-S. 2022.
\newblock Few-shot adaptation of pre-trained networks for domain shift.
\newblock In \emph{IJCAI}.

\bibitem[{Zhou et~al.(2021)Zhou, Yang, Qiao, and Xiang}]{zhou2021mixstyle}
Zhou, K.; Yang, Y.; Qiao, Y.; and Xiang, T. 2021.
\newblock Domain generalization with MixStyle.
\newblock In \emph{ICLR}.

\end{thebibliography}
